\documentclass[10pt,twocolumn,letterpaper]{article}

\usepackage{cvpr}              % To produce the CAMERA-READY version
\definecolor{cvprblue}{rgb}{0.21,0.49,0.74}
\usepackage[pagebackref,breaklinks,colorlinks,allcolors=cvprblue]{hyperref}

\def\paperID{*****} % *** Enter the Paper ID here
\def\confName{CVPR}
\def\confYear{2026}

\title{Technical Report of RoboSpatial Challenge at CVPR 2026:\\Selective Reasoning Activation and Reference-Frame Disambiguation for Embodied Spatial Reasoning}

\author{
Yuxiang Xie$^{1}$ \quad Qi Lv$^{1}$ \quad Jianming Xing$^{1}$ \quad Zijian Hong$^{1}$\\
Xiang Deng$^{1,2,*}$ \quad Weili Guan$^{1}$ \quad Liqiang Nie$^{1}$\\
$^{1}$Harbin Institute of Technology, Shenzhen,  $^{2}$Ruoyu Technology\\
$^{*}$Corresponding author\\
{\tt\small 25B951074@stu.hit.edu.cn}
}

\begin{document}
\maketitle
\begin{abstract}
Vision-language models achieve strong general perception but often struggle with the spatial reasoning required for embodied tasks. We present RoboSpatialBrain, our submission to the RoboSpatial Challenge at the Embodied Reasoning in Action Workshop, CVPR 2026, built on RoboBrain2.5-8B-NV. RoboSpatialBrain combines two training-free, inference-time mechanisms: a forced \texttt{<think>} prefix activation strategy paired with a task-specific post-prompt that elicits deliberate reasoning on context and compatibility tasks, and an explicit reference-frame redirection pipeline that resolves camera-centric and object-centric ambiguity for context tasks. We additionally explore fine-tuning RoboBrain2.5 on compatibility data and present a detailed analysis of its interaction with prompting. RoboSpatialBrain achieved first place in the RoboSpatial Challenge, with an overall success rate of 80.9\% on RoboSpatial-Home. Code is available at \url{https://github.com/YuxiangXie2003/RoboSpatialBrain}.
\end{abstract}    
\section{Introduction}
\label{sec:intro}

Spatial reasoning is a fundamental capability for embodied AI systems operating in real-world environments. While vision-language models (VLMs) achieve strong performance on tasks such as object detection and image captioning, prior work has shown that they consistently struggle with basic spatial relations~\cite{kamath2023whatsup} and often require specialized training to acquire reliable spatial reasoning abilities~\cite{chen2024spatialvlm}. A further source of difficulty is reference-frame ambiguity: VLMs tend to default to a camera-centric interpretation of spatial language and fail to generalize when a query instead requires an object-centric frame of reference~\cite{li2025viewspatialbench}.
 
The RoboSpatial-Home benchmark~\cite{song2025robospatial} evaluates these abilities through three task types. \textbf{Context} tasks ask the model to identify vacant space in a specified direction relative to a reference object. \textbf{Compatibility} tasks additionally require assessing whether the identified space can accommodate a target object given its metric dimensions. \textbf{Configuration} tasks require inferring relative positional relationships between two objects.
 
In this report, we describe \textbf{RoboSpatialBrain}, our submission to the RoboSpatial Challenge at the Embodied Reasoning in Action (ERA) Workshop, CVPR 2026. We build on RoboBrain2.5-8B-NV~\cite{tan2026robobrain25depthsight}, chosen because it is pre-trained on embodied-task-specific spatial corpora such as 2D/3D grounding, inter-object spatial relationships, and depth-conditioned localization, which directly aligns it with the demands of the RoboSpatial benchmark. We extend it with two inference-time mechanisms: prompt-based reasoning activation for reasoning-required tasks, and an explicit reference-frame redirection pipeline for context tasks. We also explored fine-tuning the model for the compatibility task, with the corresponding analysis presented in Section~\ref{sec:method-finetune} and Section~\ref{sec:exp-finetune}. Our approach achieves an overall success rate of 80.9\% on the RoboSpatial Challenge.

\section{Methodology}
\label{sec:methodology}

\begin{figure*}[t]
  \centering
  \includegraphics[width=0.9\textwidth]{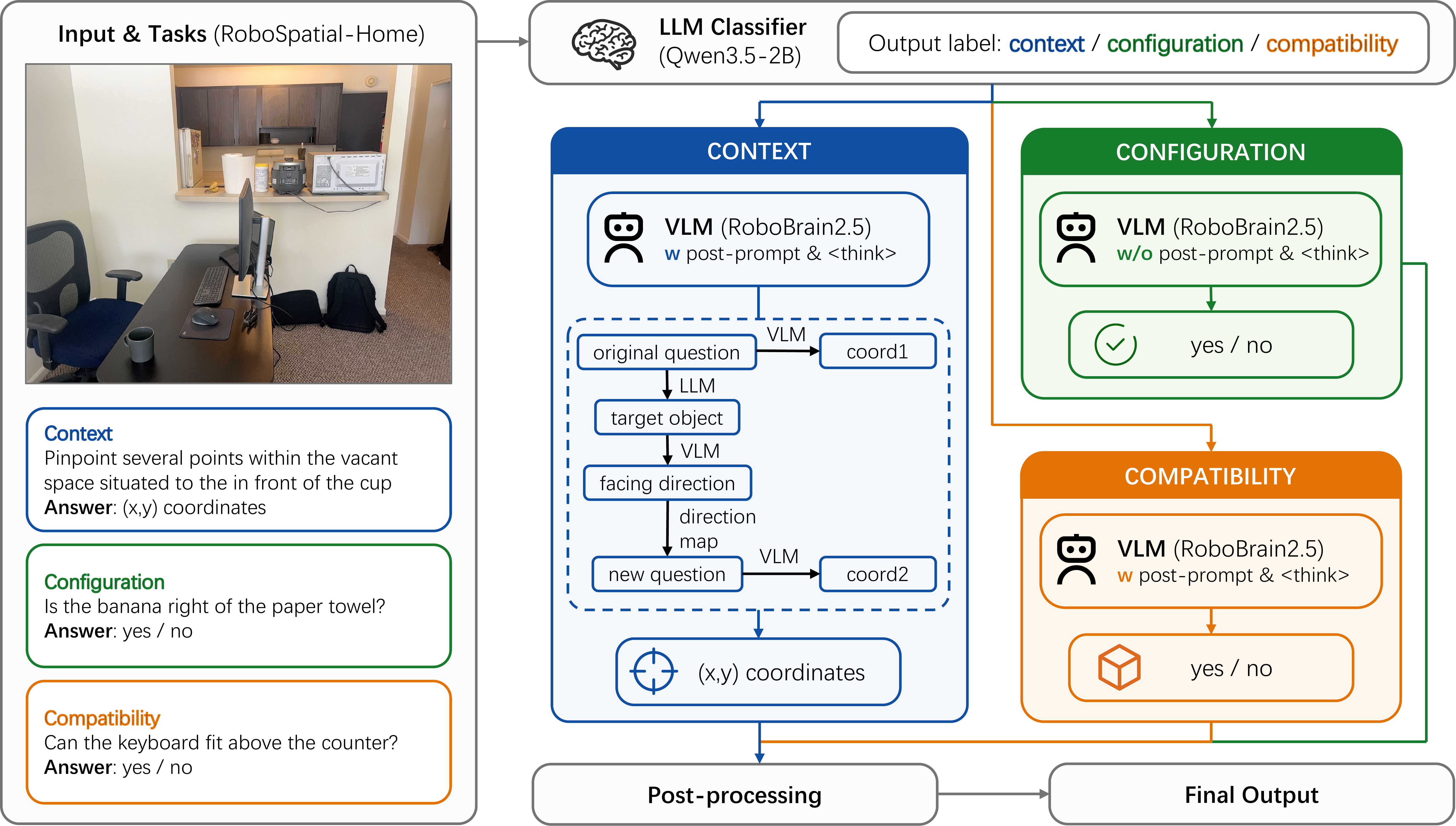}
  \caption{Overview of the RoboSpatialBrain inference pipeline. Given an input image and a spatial query, the query is first routed by task category. For \emph{context} and \emph{compatibility} queries, a post-prompt is injected and a \texttt{<think>} prefix is forced onto the model's output to elicit deliberate reasoning before the final answer is produced. For \emph{context} queries specifically, an additional reference-frame redirection module disambiguates between camera-centric and object-centric interpretations of the query before the final answer is generated. \emph{Configuration} queries are passed to the base model without modification.}
  \label{fig:overview}
\end{figure*}

Figure~\ref{fig:overview} gives an overview of the RoboSpatialBrain inference pipeline. We describe each component below.
 
\subsection{Thinking Activation and Prompting}
 
We observe that compatibility and context tasks demand a higher degree of reasoning ability from the model than configuration tasks do. Context tasks require understanding what constitutes vacant space and reasoning about which regions qualify as usable vacant space, while compatibility tasks additionally require the model to judge, alongside the existence of vacant space, whether that space is large enough to accommodate the target object. We group these two task types together as \emph{reasoning-required} tasks. Configuration tasks, in contrast, only require directly outputting the relative positional relationship between two target objects, and we group this task type as \emph{reasoning-free}.
 
An intuitive idea is to explicitly induce reasoning in the model to improve success rates on reasoning-required tasks. Since RoboBrain2.5 does not expose a standard interface for explicitly toggling a thinking mode, we instead force a \texttt{<think>} prefix onto the model's output. We additionally inject a post-prompt, detailed in Appendix~\ref{app:prompts}. Together, the forced \texttt{<think>} prefix and the post-prompt guide the model's reasoning process. This intervention improves performance on reasoning-required tasks while reducing performance on the reasoning-free task. Our final setting therefore applies both the forced \texttt{<think>} prefix and the post-prompt for compatibility and context queries, and applies no additional intervention for configuration queries. The corresponding ablation and analysis are presented in Section~\ref{sec:exp-prompt}.

\begin{figure}[t]
  \centering
  \includegraphics[width=0.9\linewidth]{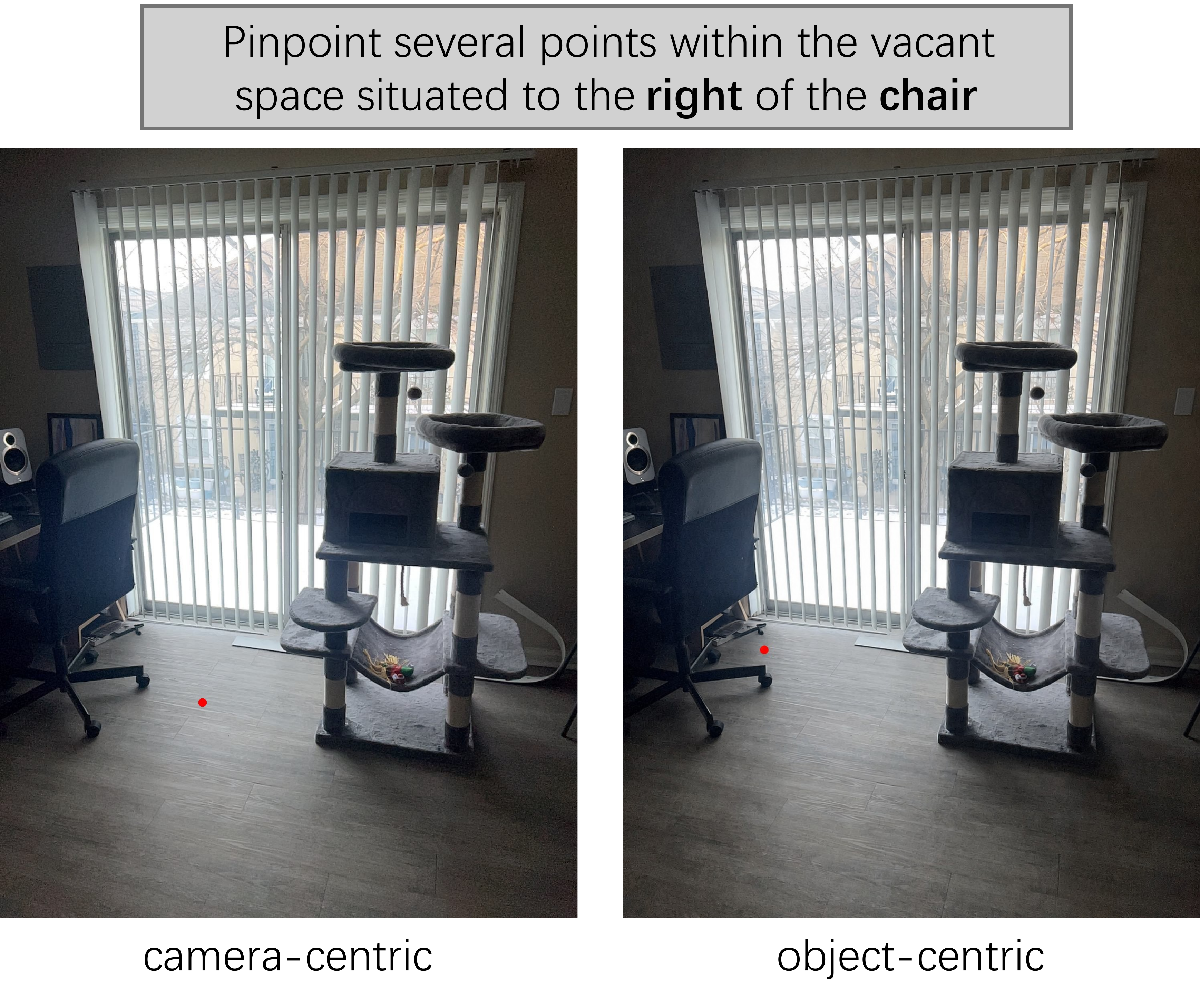}
  \caption{Camera-centric versus object-centric interpretations of the same context-task prompt. Many objects, such as chairs, acquire a canonical facing direction because a person seated in the object naturally imposes their own forward-facing direction onto it, which is why directional language describing such objects is especially prone to this camera-centric and object-centric ambiguity.}
  \label{fig:direction}
\end{figure}

\subsection{Reference Frame Redirection for Context Tasks}
 
We empirically observe that RoboBrain2.5 tends to produce camera-centric spatial predictions by default. In practice, however, people frequently describe spatial relationships in an image using an object-centric frame of reference instead. Figure~\ref{fig:direction} illustrates this ambiguity using the context-task prompt ``Pinpoint several points within the vacant space situated to the right of the chair'': depending on whether ``right'' is interpreted relative to the camera or relative to the chair's own facing direction, the predicted points land in entirely different regions of the image.

We first attempted to resolve this ambiguity purely through prompting: we informed the model of the mapping relationship between object-centric and camera-centric directions and asked it to perform the conversion itself before producing a final answer. However, RoboBrain2.5 does not reliably follow this instruction.
 
We therefore design an explicit conversion pipeline instead. We use a lightweight language model, Qwen3.5-2B~\cite{qwen35blog}, to extract the target object referenced in the query. RoboBrain2.5 is then prompted to determine the facing direction of the target object. Using a precomputed direction mapping table, the original object-centric query is rewritten into its camera-centric equivalent, and RoboBrain2.5 is queried again on this rewritten version. The final output is obtained by merging the answer to the original query with the answer to the converted query. This pipeline requires no model fine-tuning. The corresponding ablation and analysis are presented in Section~\ref{sec:exp-redirect}.
 
\subsection{Fine-Tuning for the Compatibility Task}
\label{sec:method-finetune}
 
In addition to the inference-time mechanisms above, we explored fine-tuning RoboBrain2.5 on supplementary compatibility data constructed from EmbodiedScan~\cite{embodiedscan}. Under the most basic setting, without the post-prompt and without the forced \texttt{<think>} prefix, fine-tuning brings a substantial performance improvement over the base model. However, under our best-performing prompting configuration, the base model still outperforms the fine-tuned checkpoints. We therefore did not include fine-tuning as a component of our final pipeline. The full setup, data construction, and detailed results are presented in Section~\ref{sec:exp-finetune}.
\section{Experiment}

\subsection{Thinking Activation and Prompting}
\label{sec:exp-prompt}
 
We compare the performance of the configuration, compatibility, and context tasks under different combinations of the post-prompt and the forced \texttt{<think>} prefix. Results are shown in Table~\ref{tab:prompt}. Both compatibility and context benefit most from jointly applying the post-prompt and the \texttt{<think>} prefix, while configuration achieves its best performance when neither intervention is applied. The compatibility task shows by far the largest gain among the three task types: jointly applying the post-prompt and the \texttt{<think>} prefix raises its success rate from 40.0\% to 83.8\%, a 43.8 percentage-point improvement.
 
We attribute this pattern to the difference in reasoning demand across task types. Although RoboBrain2.5 does not natively support chain-of-thought reasoning, forcing the \texttt{<think>} prefix nonetheless appears to have a positive effect on the model's reasoning behavior, even though no visible reasoning trace is produced. We hypothesize that, because RoboBrain2.5 is built on the Qwen3-VL architecture~\cite{Qwen3-VL}, the forced prefix implicitly activates residual reasoning-format priors carried over from the Qwen series of models. Supervised fine-tuning suppresses the model's decoding behavior so that it does not explicitly emit a reasoning trace; however, the corresponding thinking pathway may still participate in representation construction during the forward pass. This manifests as a performance improvement on tasks that require complex reasoning, and a performance decrease on simpler tasks due to overthinking.
 
Based on these results, our final configuration applies both the post-prompt and the forced \texttt{<think>} prefix for compatibility and context tasks, and applies neither for configuration tasks, yielding the row labeled ``Best Combination'' in Table~\ref{tab:prompt}.
 
\begin{table}[t]
  \centering
  \caption{Success rate (\%) on the three RoboSpatial task types under different combinations of the post-prompt and the forced \texttt{<think>} prefix. Avg. denotes the official average score, computed using the official task distribution rather than a uniform macro-average over task types. The largest value in each column is bolded, except in the final row, which reports our selected per-task-optimal combination.}
  \label{tab:prompt}
  \resizebox{\columnwidth}{!}{%
  \begin{tabular}{lcccc}
    \toprule
     & Config. & Compat. & Context & Avg. \\
    \midrule
    Post-prompt + \texttt{<think>} & 83.7 & \textbf{83.8} & \textbf{70.5} & 79.1 \\
    \texttt{<think>} only & 84.6 & 59.0 & 63.1 & 69.4 \\
    Post-prompt only & 86.2 & 64.8 & 68.8 & 73.7 \\
    Neither & \textbf{88.6} & 40.0 & 64.8 & 65.7 \\
    \midrule
    \textbf{Best Combination} & \textbf{88.6} & \textbf{83.8} & \textbf{70.5} & \textbf{80.9} \\
    \bottomrule
  \end{tabular}}
\end{table}
 
\subsection{Reference Frame Redirection}
\label{sec:exp-redirect}
 
We compare context-task performance with and without reference frame redirection. Results are shown in Table~\ref{tab:redirect}. Reference frame redirection brings a consistent performance improvement across all post-prompt and \texttt{<think>} prefix configurations, confirming that resolving the camera-centric and object-centric ambiguity is beneficial regardless of the reasoning-elicitation setting in use.
 
\begin{table}[t]
  \centering
  \caption{Context-task success rate (\%) with and without reference frame redirection, denoted w and w/o respectively. $\Delta$ denotes the gain from redirection.}
  \label{tab:redirect}
  \resizebox{\columnwidth}{!}{%
  \begin{tabular}{lccc}
    \toprule
     & \textbf{w redirect} & w/o redirect & $\Delta$ \\
    \midrule
    Post-prompt + \texttt{<think>} & \textbf{70.5} & 68.8 & $+$1.7 \\
    \texttt{<think>} only & \textbf{63.1} & 59.8 & $+$3.3 \\
    Post-prompt only & \textbf{68.8} & 66.4 & $+$2.4 \\
    Neither & \textbf{64.8} & 61.5 & $+$3.3 \\
    \bottomrule
  \end{tabular}}
\end{table}
 
\begin{table*}[t!]
  \centering
  \small
  \caption{Compatibility-task success rate (\%) for the base model and two fine-tuned checkpoints at 1{,}000 and 3{,}000 training steps, under different post-prompt and \texttt{<think>} prefix combinations. $\Delta$ denotes the change relative to the base model and is bolded.}
  \label{tab:finetuning}
  \begin{tabular}{lccccc}
    \toprule
     & Base & 1000 steps & $\Delta$(1000 steps) & 3000 steps & $\Delta$(3000 steps) \\
    \midrule
    Post-prompt + \texttt{<think>} & 83.8 & 71.4 & \textbf{$-$12.4} & 61.0 & \textbf{$-$22.8} \\
    \texttt{<think>} only & 59.0 & 61.9 & \textbf{$+$2.9} & 53.3 & \textbf{$-$5.7} \\
    Post-prompt only & 64.8 & 74.3 & \textbf{$+$9.5} & 64.8 & \textbf{$0.0$} \\
    Neither & 40.0 & 67.6 & \textbf{$+$27.6} & 61.0 & \textbf{$+$21.0} \\
    \bottomrule
  \end{tabular}
\end{table*}
 
\subsection{Fine-Tuning}
\label{sec:exp-finetune}
 
\paragraph{Training instability with mixed task data.} At the outset of this work, we attempted to improve model performance purely through fine-tuning. We first used the RoboSpatial pipeline to separately construct configuration, compatibility, and context training data, shuffled them together, and trained on the mixture. This setup caused the model to collapse: it converged to a degenerate state in which it produced a single fixed type of output per task category regardless of the input, with configuration predictions converging entirely to ``no'', compatibility predictions converging entirely to ``yes'', and context predictions converging to a small cluster of points concentrated in one region of the image.
 
\paragraph{Compatibility-only data construction.} To avoid this collapse, we restricted training to compatibility data only. We filtered the EmbodiedScan dataset~\cite{embodiedscan} to retain images that are not excessively blurry and in which the target object is clearly identifiable. From the filtered images, we constructed 24{,}000 examples in compatibility-task VQA format using the RoboSpatial data construction pipeline~\cite{song2025robospatial}. Training hyperparameters are listed in Appendix~\ref{app:hyperparams}. We evaluate checkpoints at 1{,}000 and 3{,}000 steps, selected based on a held-out validation split.
 
\paragraph{Results.} Table~\ref{tab:finetuning} compares the fine-tuned checkpoints against the base model across the same post-prompt and \texttt{<think>} prefix combinations used above. Under the most basic setting, with neither the post-prompt nor the \texttt{<think>} prefix applied, the 1{,}000-step checkpoint improves over the base model by 27.6 percentage points, and the 3{,}000-step checkpoint improves over it by 21.0 percentage points. When only the post-prompt is used, the fine-tuned model remains no worse than the base model. However, under our best-performing combination, the post-prompt together with the \texttt{<think>} prefix, the base model outperforms both fine-tuned checkpoints: the 1{,}000-step checkpoint falls 12.4 percentage points below the base model, and the 3{,}000-step checkpoint falls 22.8 percentage points below it. The same pattern holds when only the \texttt{<think>} prefix is used, and the gap widens as training proceeds from 1{,}000 to 3{,}000 steps.
 
We attribute this pattern to the simplicity of our data construction procedure: because the compatibility training data was constructed in a limited-diversity construction pipeline and covers a narrow distribution of scenes and question types, fine-tuning on it appears to degrade the model's broader reasoning and instruction-following ability, consistent with prior findings that narrow-domain supervised fine-tuning can induce catastrophic forgetting of a model's general capabilities~\cite{luo2023empirical}. This would in turn diminish the benefit that the post-prompt and the \texttt{<think>} prefix would otherwise provide. This finding is consistent with our decision to exclude fine-tuning from the final RoboSpatialBrain pipeline, and instead motivates the data-quality directions discussed in Section~\ref{sec:future-work}.
\section{Conclusion and Future Work}
\label{sec:future-work}
 
We present \textbf{RoboSpatialBrain}, the first-place solution to the RoboSpatial Challenge at the ERA Workshop, CVPR 2026, achieving an overall success rate of 80.9\%. Built on RoboBrain2.5-8B-NV, our method combines selective \texttt{<think>} prefix activation for reasoning-required tasks, namely compatibility and context, with an explicit reference-frame redirection pipeline that resolves the camera-centric and object-centric ambiguity inherent to context-task queries. We additionally explored fine-tuning RoboBrain2.5 on compatibility data and presented a detailed analysis of its effect under different prompting configurations, offering insight into how fine-tuning interacts with prompt-based reasoning elicitation.
 
For future work, the training data used in our fine-tuning experiments was drawn from a single source, EmbodiedScan, and constructed with a relatively simple pipeline. Broadening both the data sources and the diversity of scenes, object categories, and question phrasings, in the spirit of the data construction strategies used to train RoboBrain2.5 itself~\cite{tan2026robobrain25depthsight}, may help fine-tuning better preserve the model's general reasoning ability while still improving task-specific performance. Higher-quality 3D scene datasets such as ScanNet++~\cite{yeshwanth2023scannet++} could also serve as a complementary data source going forward.
{
    \small
    \bibliographystyle{ieeenat_fullname}
    \bibliography{main}
}

\clearpage
\setcounter{page}{1}
\maketitlesupplementary

\appendix

\begin{table}[t]
  \centering
  \caption{Fine-tuning hyperparameters.}
  \label{tab:hyperparams}
  \begin{tabular}{ll}
    \toprule
    Hyperparameter & Value \\
    \midrule
    LoRA rank & 16 \\
    GPUs & 4 $\times$ NVIDIA A100 \\
    Global batch size & 4 \\
    Optimizer & AdamW \\
    Learning rate & $1\times10^{-4}$ \\
    LR decay floor & $1\times10^{-5}$ \\
    Total steps & 4{,}000 \\
    \bottomrule
  \end{tabular}
\end{table}

\section{Post-Prompts}
\label{app:prompts}

We list below the post-prompts used for each task type when the post-prompt intervention is applied, as described in Section~\ref{sec:exp-prompt}. The configuration prompt was used only in ablation experiments and is not used in the final pipeline.
 
\paragraph{Configuration.}
\begin{quote}
\small
\texttt{Your task is to answer the question above. Respond with a brief explanation if needed, followed by a yes or no answer in the last line of your response.}\\
\texttt{Format your final answer strictly as follows on the last line of your response:}\\
\texttt{Answer: yes or no}\\
\texttt{Do not include additional text after this line}
\end{quote}
 
\paragraph{Compatibility.}
\begin{quote}
\small
\texttt{Look at the spatial positions and relationships of objects in the image. Based on your visual analysis, answer yes or no.}\\
\texttt{Your final line must be: Answer: yes or no}
\end{quote}
 
\paragraph{Context.}
\begin{quote}
\small
\texttt{You MUST provide at least 5 distinct 2D points that satisfy the conditions above.}\\
\texttt{Do NOT output only one point. Format your final answer strictly as a list of tuples:}\\
\texttt{[(x1, y1), (x2, y2), (x3, y3), ...].}
\end{quote}

\section{Fine-Tuning Hyperparameters}
\label{app:hyperparams}
 
Table~\ref{tab:hyperparams} lists the hyperparameters used for the compatibility-task fine-tuning experiment described in Section~\ref{sec:method-finetune}.

\end{document}